\documentclass[letterpaper]{article} %
\usepackage{aaai23}  %
\usepackage{times}  %
\usepackage{helvet}  %
\usepackage{courier}  %
\usepackage[hyphens]{url}  %
\usepackage{graphicx} %
\usepackage{natbib}  %
\usepackage{caption} %
\usepackage{subcaption}
\usepackage{algorithm}
\usepackage{algorithmic}

\usepackage{newfloat}
\usepackage{listings}
\usepackage{makecell}
\usepackage{amssymb}
\usepackage{amsmath}
\DeclareCaptionStyle{ruled}{labelfont=normalfont,labelsep=colon,strut=off} %
\floatstyle{ruled}
\newfloat{listing}{tb}{lst}{}
\floatname{listing}{Listing}
\title{Real-World Deep Local Motion Deblurring}
\author{
    Haoying Li\textsuperscript{\rm 1,\rm 2},
    Ziran Zhang\textsuperscript{\rm 1}\equalcontrib,
    Tingting Jiang\textsuperscript{\rm 2}\equalcontrib,\\
    Peng Luo\textsuperscript{\rm 1}\equalcontrib,
    Huajun Feng\textsuperscript{\rm 1}{\rm }\thanks{Corresponding author},
    Zhihai Xu\textsuperscript{\rm 1}
}
\affiliations{
    \textsuperscript{\rm 1} College of Optical Science and Engineering, Zhejiang University \\
    \textsuperscript{\rm 2} Research Center for Intelligent Sensing Systems, Zhejiang Laboratory \\

    \{lhaoying, naturezhanghn, luop, fenghj, xuzh\}@zju.edu.cn,
    eagerjtt@zhejianglab.com\\
 
}

\usepackage{bibentry}

\begin{document}

\maketitle

\begin{abstract}
Most existing deblurring methods focus on removing global blur caused by camera shake, while they cannot well handle local blur caused by object movements. To fill the vacancy of local deblurring in real scenes, we establish the first real local motion blur dataset (ReLoBlur), which is captured by a synchronized beam-splitting photographing system and corrected by a post-progressing pipeline. Based on ReLoBlur, we propose a Local Blur-Aware Gated network (LBAG) and several local blur-aware techniques to bridge the gap between global and local deblurring: 1) a blur detection approach based on background subtraction to localize blurred regions; 2) a gate mechanism to guide our network to focus on blurred regions; and 3) a blur-aware patch cropping strategy to address data imbalance problem. Extensive experiments prove the reliability of ReLoBlur dataset, and demonstrate that LBAG achieves better performance than state-of-the-art global deblurring methods and our proposed local blur-aware techniques are effective.
\end{abstract}

\section{Introduction} \label{Intro}
Single image deblurring has been persistently analyzed \cite{wang2017deepdeblur,zhou2019davanet,jin2021multi,zhou2022lednet} and motion blur can be categorized into two kinds: global motion blur and local motion blur. In a global motion blurred image, blur exists in all regions of the image, and is usually caused by camera shake \cite{schelten2014localized,zhang2018learning}. Fantastic progress has been made in global motion deburring \cite{nah2017deep,Kupyn_2019_ICCV,chen2021hinet}. However, local motion deblurring is under few explorations, where blurs only exist in some of the regions of the image, and are mostly caused by object movements captured by a static camera.

Deep local motion deblurring in real scenes is a vital task with many challenges. Firstly, there is no public real local motion blur dataset for deep learning. Secondly, local motion deblurring is a complicated inverse problem due to the random localization of local blurs and the unknown blur extents. Besides, the blurred regions occupy only a small proportion of the full image, causing a deep neural network to pay too much attention to the background. This data imbalance issue is contrary to the goal of local deblurring. 

To tackle the local motion blur problems, data is the foundation. Existing deblurring datasets \cite{nayar2004motion,kohler2012recording,hu2016image,nah2017deep} are mainly constructed for global deblurring with camera motion. Among them, a widely used approach is synthesizing blurred images by convolving either uniform or non-uniform blur kernels with sharp images \cite{boracchi2012modeling,schuler2015learning, chakrabarti2016neural}. However, this approach cannot assure the fidelity of blurred images. Another kind of approach shakes the camera to mimic blur caused by camera trembling, and averages consecutive short-exposure frames to synthesize global blurred images \cite{nah2017deep,nah2019ntire}. However, this kind of averaging approach cannot simulate real overexposure outliers due to the dynamic range of each frame \cite{chang2021beyond}. In this paper, we establish a real local motion blur dataset, ReLoBlur, captured by a static synchronized beam-splitting photographing system, which can capture local blurred and sharp images simultaneously. ReLoBlur only contains blur caused by moving objects, without camera motion blur. Moreover, we propose a novel paired image post-processing pipeline to address color cast and misalignment problems occurred in common beam-splitting systems \cite{tai2008image,rim2020real}. To the best of our acknowledgment, ReLoBlur is the first real local motion blur dataset captured in nature.

Since deep learning is proven to be successful in the global deblurring task, a direct way to remove local blur is to borrow ideas from deep global deblurring models. However, unlike global motion blur, local motion blur changes abruptly rather than smoothly at object boundaries, and the sharp background remains clear \cite{schelten2014localized}. Therefore, global deblurring networks fail to handle local motion blur and may raise image artifacts in sharp backgrounds. Enlightened by MIMO-Unet \cite{cho2021rethinking}, we propose a Local Blur-Aware Gated deblurring method (LBAG) which localizes blurred regions and predicts sharp images simultaneously with novel local blur-aware deblurring techniques. To localize blurred regions, LBAG is trained to predict local blur masks under the supervision of local blur mask ground-truths, which are generated by a blur detection approach based on background subtraction. With the help of predicted local blur masks, we safely introduce a gate block to LBAG, which guides the network to focus on blurred regions. To address the data imbalance issue, we propose a Blur-Aware Patch Cropping strategy (BAPC) to assure at least 50\% of the input training patches contain local blur. We conduct experiments on ReLoBlur dataset and evaluate our proposed method in terms of PSNR, SSIM, aligned PSNR \cite{rim2020real}, weighted PSNR \cite{jamali2021weighted} and weighted SSIM, the latter two of which specifically measure local performances. To sum up, our main contributions are:
\begin{itemize}
    \item We establish the first real local motion blur dataset, ReLoBlur, captured by a synchronized beam-splitting photographing system in daily real scenes, and corrected by a post-processing pipeline. ReLoBlur contains 2405 image pairs and we will release the dataset soon.
    \item We develop a novel local blur-aware gated network, LBAG, with several local blur-aware techniques to bridge the gap between global and local deblurring.
    \item Extensive experiments show that ReLoBlur dataset enables efficient training and vigorous evaluation, and the proposed LBAG network exceeds other SOTA deblurring baselines quantitatively and perceptually.
\end{itemize}

\section{Related Works}\label{Section:related_works}
\subsection{Image Deblurring Datasets}
Local motion blur dataset draws little attention, while global blur datasets update rapidly. Approaches for generating global blurred images include: 1) convolving sharp images blur kernels \cite{chakrabarti2016neural, schuler2015learning, sun2015learning}; 2) averaging consecutive frames at very short intervals and selecting the central frame as the sharp image \cite{nah2017deep, nah2019ntire}; 3) using a coaxial beam-splitting system to simultaneously capture global blurred-sharp image pairs \cite{rim2020real}. 
Approach 1 and 2 are not suitable to capture real blur, as illustrated in \ref{Intro}. Approach 3 has drawbacks in color cast which is obvious in RealBlur dataset \cite{rim2020real}. This should be avoided because it may reduce the training performance in local deblurring tasks (We discuss this in our project's homepage\footnote{https://leiali.github.io/ReLoBlur\_homepage/index.html\label{web}}).
To the best of our acknowledgment, there is no public dataset containing real local blur. We establish the first real local motion blur dataset, ReLoBlur, captured by a simultaneous photographing system and corrected through our post-processing pipeline. 

\subsection{Single Image Deep Deblurring Methods}
Single image deep deblurring methods mainly focus on global blur tasks. Nah et al. \cite{nah2017deep} introduced a multi-scale convolutional neural network that restores sharp images in an end-to-end manner. SRN-DeblurNet \cite{tao2018scale} overcame the training instability of multi-scaled deblurring networks by sharing network weights across scales. DeblurGAN \cite{kupyn2018deblurgan} enhanced perceptual image quality of global deblurring by using perceptual losses. DeblurGAN-v2 \cite{Kupyn_2019_ICCV} accelerated DeblurGAN on training speed. Chen et al. \cite{chen2021hinet} proposed HINET and refreshed the global deblurring score in the 2021 NITIRE Chanllenge. Nevertheless, these methods are not well-adopted in local deblurring tasks. Some algorithms \cite{schelten2014localized, pan2016soft} managed to deblur locally, but were limited within kernel estimation. Enlightened by Cho et al. \cite{cho2021rethinking}, we propose a local blur-aware gated deblurring method for deep local deblurring, which recovers local blurred images by implementations of a gate block, a blur-aware patch cropping strategy and a local blur foreground mask generator.

\section{ReLoBlur Dataset}
\label{Section:ReLoBlur}
In this section, we firstly describe the formation of local motion blur. Then, we introduce our image acquisition process based on the blur model. Finally, we introduce the paired image post-processing pipeline.
\subsection{The Formation of Local Motion Blur}
\label{Section:localblur_model}
During camera exposure, the moving object's actions accumulate as the camera sensor receives light every time, forming spatial aliasing, which presents local blur in real captured images. The local motion-blurred image formation can be modeled as:
\begin{equation}\small
   \mathrm{B}(x,y) = \mathrm{ISP}\left(\int_{T_1}^{T_2}f(t,x,y) dt\right),
	\label{equ_local_blur_model}
\end{equation}
where $\mathrm B$ denotes the locally blurred color image, $T_2 - T_1$ denotes the total time of an exposure, and $(x,y)$ denotes the pixel location. $f(t,x,y)$ means the photon response of pixel $(x,y)$ at a time $t$. When $T_1 \to T_2$, Eq. \ref{equ_local_blur_model} is the formation of the corresponding sharp ground-truth image.
$\mathrm{ISP}$ means image signal processing operations, generating a Bayer raw image into a colorful RGB image.

\subsection{Paired Image Acquisition}
\label{section:image_acq}
\begin{figure}[t]
\centering
\includegraphics[width=0.48\textwidth]{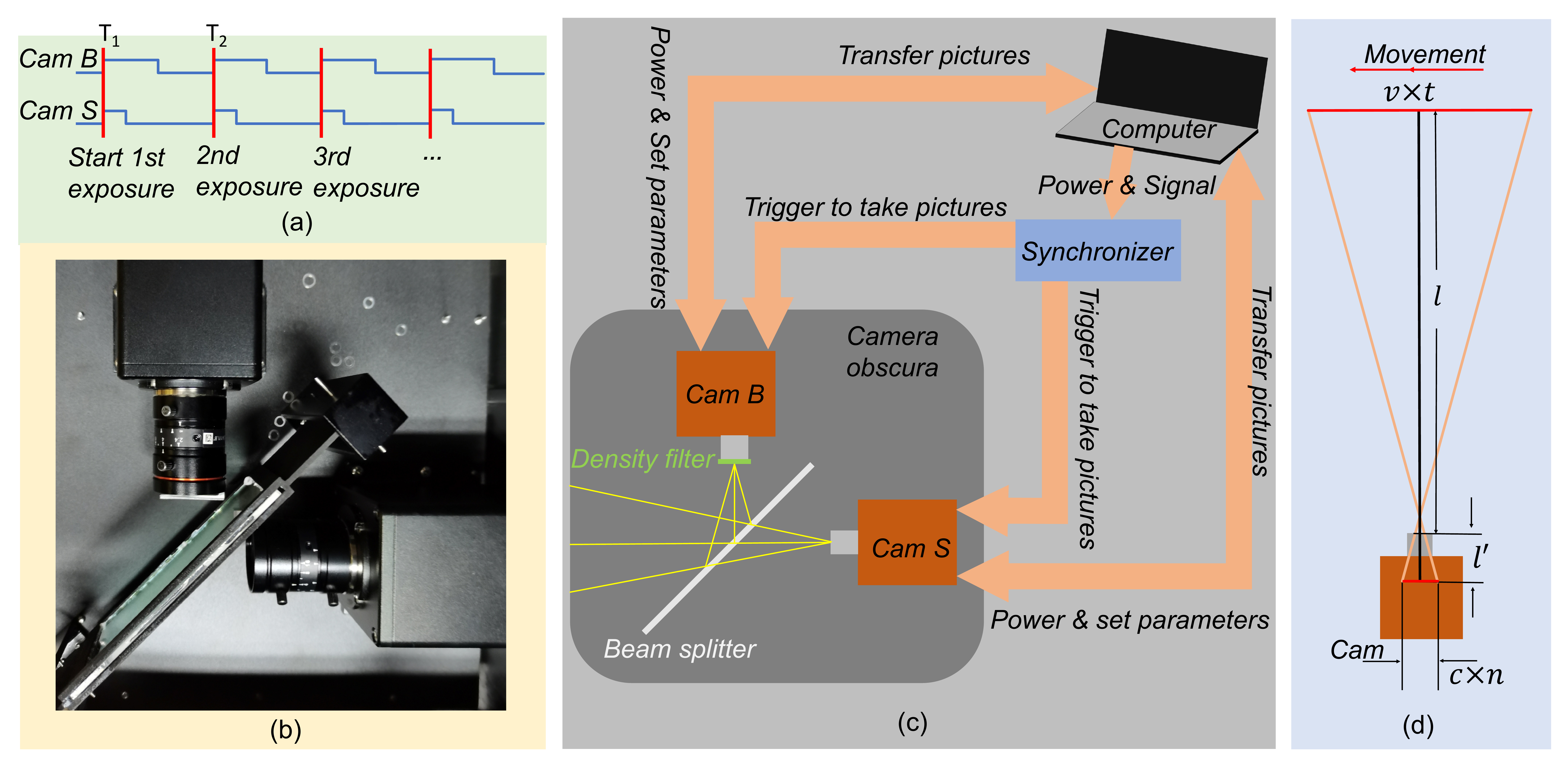}
\caption{Overview of the paired image acquisition system: (a) the exposure mode of Camera B and S following the local motion-blurred image formation; (b) a real picture of Camera B, S and the beam-splitting plate inside a camera obscure; (c) the synchronized beam-splitting photographing system; (d) the imaging model.}
\label{fig:data_aqu_sys}
\end{figure}
We manage to capture real local motion-blurred images by reproducing the formation of local motion blur mentioned above. We use a scientific camera (Camera B) to collect local blurred image in a long exposure time $t_L$ to accumulate local blur of moving objects. Simultaneously, we use another same camera (Camera S) to capture the corresponding sharp images in a very short exposure time $t_S$. The two cameras start exposing at the same time but end after different exposure times, as shown in Fig. \ref{fig:data_aqu_sys}(a). In order to share the two cameras same scene, Camera B and S are placed on the reflection end and transmission end of a beam-splitting plate, respectively. Each camera is placed 45 degrees to the 50\% beam-splitting plate (the transmittance and reflectivity per spectrum is 50\%), as shown in Fig.\ref{fig:data_aqu_sys}(b)(c). In front of Camera B is a density filter with transmittance $\tau=\frac{t_S}{t_L}$, assuring the equivalence of photon energy received by the two cameras. Both Camera B and S are connected to a synchronizer, which triggers the two cameras to start exposing simultaneously in every shot but end exposure after different times. A computer is connected to the synchronizer, Camera B and S, powers and transmits data. In this way, the beam splitting photographing system can capture locally blurred images and their corresponding sharp images simultaneously according to the formation of local motion blur.

Before capturing image pairs, we adjust the settings of the synchronized beam-splitting photographing system according to the relationship between the exposure time and the object distance:
\begin{equation}\small
	t_S = \frac{c\times n\times d}{l' \times v},
	\label{equ_capturing_rule}
\end{equation}
where $c$, $n$, $d$, $l'$, $v$ and $t_S$ denote sensor-pixel side length, desired blurry pixels, object distance, image distance, object moving speed and short exposure time of Camera S, respectively, as shown in Fig. \ref{fig:data_aqu_sys} (d). Eq. \ref{equ_capturing_rule} indicates that the short exposure time is closely related to desired blurry pixels and object distance. Hence, we adjust the exposure time according to the desired blurry pixels and object distances, and we set the long exposure times as $t_L = \frac{t_S}{\tau}$.
	
\begin{figure*}[h]
\centering
\includegraphics[width=\textwidth]{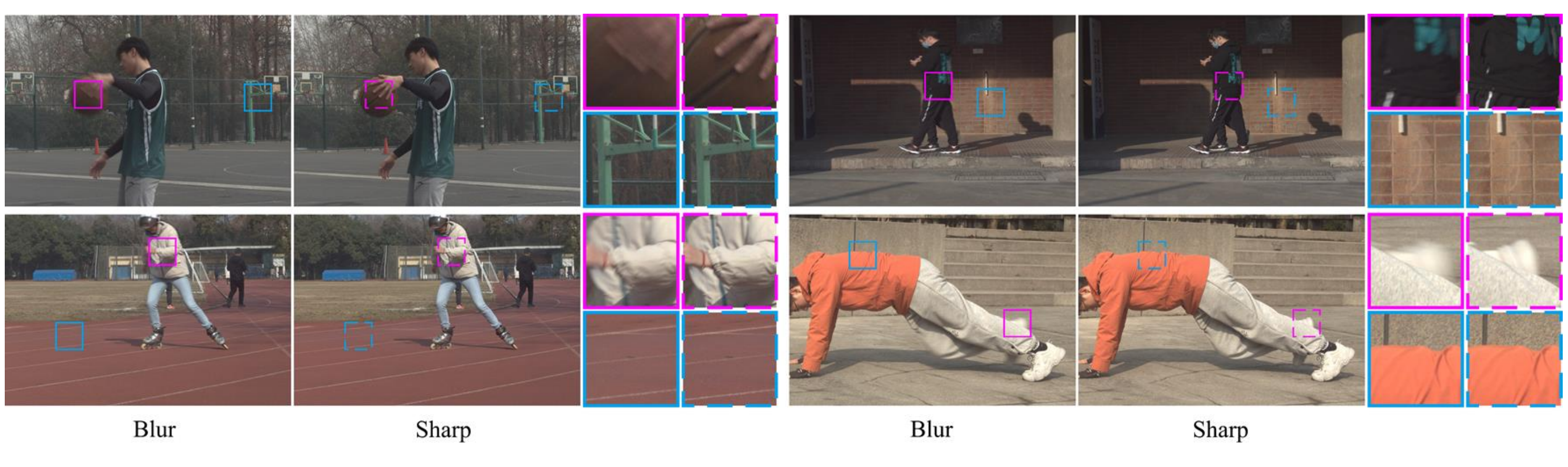}
\caption{Examples of ReLoBlur dataset: the $1^{st}$ and $5^{th}$ columns are local motion-blurred images. The $2^{nd}$ and $6^{th}$ columns are the corresponding sharp images. The pink solid boxes and pink dotted boxes denote the locally blurred regions from locally blurred images, and the corresponding sharp regions from sharp images. The blue solid boxes and the blue dotted boxes denote the sharp regions from locally blurred images and the corresponding sharp regions from sharp images.}
\label{fig:dataset}
\end{figure*}

In every scene, we capture a pair of whiteboard images for color correction, and a pair of static background images as reference for geometrical alignment in post-processing. We capture 2405 pairs of RAW images, conduct post-progressing to correct image degradations and convert RAW images to colorful RGB images with the image size of $2152\times1436$. ReLoBlur includes but is not limited to indoor and outdoor scenes of pedestrians, vehicles, parents and children, pets, balls, plants and furniture. The local motion blur sizes range from 15 pixels to 70 pixels in blurry images, and no more than 6 pixels in sharp images. On average, the blur regions take up 11.75\% of the whole image area in ReLoBlur dataset. We show our ReLoBlur dataset in Fig. \ref{fig:dataset} and our project's home page \ref{web}.

\subsection{Paired Image Post-processing}
\begin{figure*}[t]
    \centering
    \includegraphics[width=\textwidth]{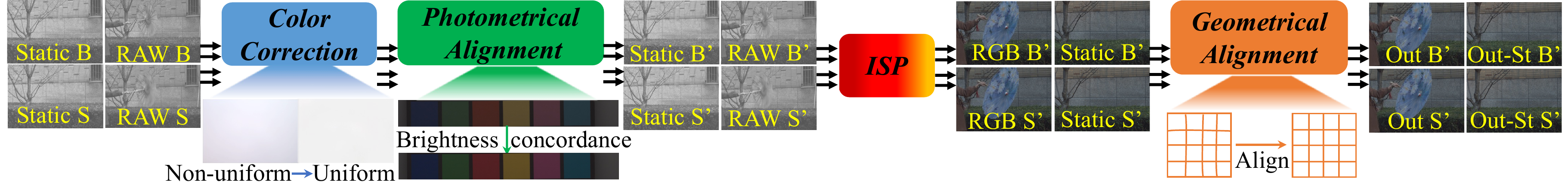}
    \caption{Paired image post-processing pipeline.}
    \label{fig:post-process}
\end{figure*}

As shown in Fig. \ref{fig:post-process}, common coaxial systems raise several image degradations: 1) physically, the beam-splitting plate's transmission-reflection ratio varies with the incident angle, causing location-oriented color cast \cite{wang2009wide,fu2010measurement}, which reduces visual perception and may worsen local deblurring performance; 2) though the cameras and lens at the transmission and reflection ends are of the same module, there still exists a slight brightness difference between a pair of blur and sharp images, due to the transmittance and reflectivity deviation of the beam-splitting plate and the unavoidable discrepancy of photovoltaic conversion of the two sensors; 3) despite carefully adjusting cameras' locations, spatial misalignments still exist because of unavoidable mechanical error. In this paper, we design a paired image post-processing pipeline to correct the above problems. The post-processing pipeline includes color correction, photometrical alignment, ISP and geometric alignment, as shown in Fig. \ref{fig:post-process}.

\subsubsection{Color Correction}
To solve the location-oriented color cast problem, we apply color correction coefficients $\alpha$ to conduct color correction on Bayer RAW images:
\begin{equation}\small
   [P'_{R_k}, P'_{G_k}, P'_{B_k}] = [\alpha_{R_k}, \alpha_{G_k}, \alpha_{B_k}] \cdot [P_{R_k}, P_{G_k}, P_{B_k}],
	\label{equ_correct_color}
\end{equation}
where $P$ and $P'$ denote RAW pixel values of a local patch in different color channels before and after color correction, respectively. $k\in\{1,2,3,...,K\}$ is the patch number and $K$ is the total patch number. The color correction coefficient $\alpha$ could be obtained by multiplying pixel coordinates and location-oriented color constants $\{a_0, a_1, a_2, ..., a_9\}$:
\begin{align}\small
    	\begin{bmatrix}
\alpha_{R_1}&\cdots&\alpha_{R_k}&\cdots&\alpha_{R_K}\\
        	\alpha_{G_1}&\cdots&\alpha_{G_k}&\cdots&\alpha_{G_K}\\
        	\alpha_{B_1}&\cdots&\alpha_{B_k}&\cdots&\alpha_{B_K}
        \end{bmatrix}
    	= 
    	 \begin{bmatrix}
        	1 {~} 1 {~} 1
        	\end{bmatrix}^T
        \times\\\nonumber
        \begin{bmatrix}
    	    a_0 & a_1 &\cdots& a_9\\
    	\end{bmatrix}
        \times
        Z,
\end{align}
\begin{equation}\small
        Z = \begin{bmatrix}
        	{x_1}^3 & {x_1}^2{y_1} & {x_1}{y_1}^2 & \dots & {y_1}^2 & {x_1}        & {y_1}        & 1 \\
        	\vdots  &              &              & \ddots  &     &    &     
        	        &                             \vdots \\
        	{x_K}^3 & {x_K}^2{y_K} & {x_K}{y_K}^2 &\dots & {y_K}^2 & {x_K}        & {y_K}        & 1 
        \end{bmatrix}^T
    \label{equ_color_calibration}
\end{equation}

The location-oriented color constants vary with different cameras. We manage to fix the location-oriented color constants by single camera color calibration in the laboratory. 

For each camera, we capture a RAW image of a standard 6500K transmissive lightbox in a darkroom, and divide a RAW image into $k$ patches. Each patch's location is its central pixel location $(x_k,y_k)$. Because there is no obvious color cast at the central region of each camera, we choose the central patch of each camera as the target patch. We calculate $\alpha$ of each patch $k$ by using Eq. \ref{equ_correct_color}, replacing $P'$ with the average pixel value of the patch $k$, and $P$ with the average pixel value of the central patch. After obtaining $\alpha$ of each path by channels, the location-oriented color constants are fixed by inversely deviating Eq. \ref{equ_color_calibration}.
In real capturing, we obtain color correction coefficient $\alpha$ applying Eq. \ref{equ_color_calibration} pixel by pixel and correct the color cast on RAW images by Eq.\ref{equ_correct_color}.

\subsubsection{Photometrical Alignment}
 To eliminate the brightness difference between the sharp and locally blurred images, we photometrically align the locally blurred images to their corresponding sharp images through every color channel. Different from \cite{rim2020real}, we adjust the image brightness by parameterizing a brightness correction coefficient $\beta$:
\begin{equation}\small
        \beta = (\beta_{\textit{R}}, \beta_{\textit{G}}, \beta_{\textit{B}}) =(\frac{\overline {P_{R_\mathrm{S}}}}{\overline {P_{R_\mathrm{B}}}}, 
        \frac{\overline {P_{G_\mathrm{S}}}}{\overline {P_{G_\mathrm{B}}}},
        \frac{\overline {P_{B_\mathrm{S}}}}{\overline {P_{B_\mathrm{B}}}}),
	\label{equ_photometric_algign}
	\end{equation}
where $\overline P$ denotes the channel mean value of a blur (or sharp) image. $R$, $G$ and $B$ stand for red, green and blue channels, respectively. $\mathrm{B}$ and $\mathrm{S}$ denote blur and sharp image, respectively.
Set $P_{\mathrm B}$ and $P'_{\mathrm B}$ to be the pixel value of a locally blurred image before and after photometrical alignment, respectively. For each pixel at location $(x,y)$, a locally blurred image captured by Camera B is photometrically aligned to:
\begin{equation}\small
    P'_\mathrm{B}(x,y) = \beta \cdot P_\mathrm{B}(x,y).
	\label{photometrical}
	\end{equation}
	
\subsubsection{ISP Operations}
 After photometrical alignment, we generate RGB images from RAW images by sequential ISP operations, as described in Eq. \ref{equ_local_blur_model}. We firstly demosaic the RAW images using Menon's method \cite{menon2006demosaicing}. Then we conduct white-balancing, color-mapping and Gamma correction in turns. 
 
\subsubsection{Geometrical Alignment}
We geometrically align locally blurred images to their corresponding sharp images by estimating optical flow and conduct image interpolation. In order to align the static background in every scene, we firstly use Camera B and S to capture a pair of static images as a reference image pair. While capturing, the photographing system, the background and foreground objects all remain still. Thus, the reference image pair can be used to calibrate the optical flow between the two cameras. In calibration, we use \textit{Pyflow} method \cite{pathak2017learning} to calculate the optical flow. In the reference image pair, because the foreground object stands at the same location where it will move, there is almost no depth difference. The computed optical flow can be generally applied to all blurred-sharp image pairs in the same scene. To better maintain edges, we utilize CMTF method \cite{freedman2016constant} for interpolation. 

To prove the authenticity of ReLoBlur dataset, we quantitatively and perceptually evaluate the results of our paired image post-processing pipeline, including color correction, photometrical alignments and geometrical alignments. Details are shown in our project's home page \ref{web}.

\section{Local Blur-Aware Gated Model}
\begin{figure*}[t]
	\centering
	\includegraphics[width=0.9\textwidth]{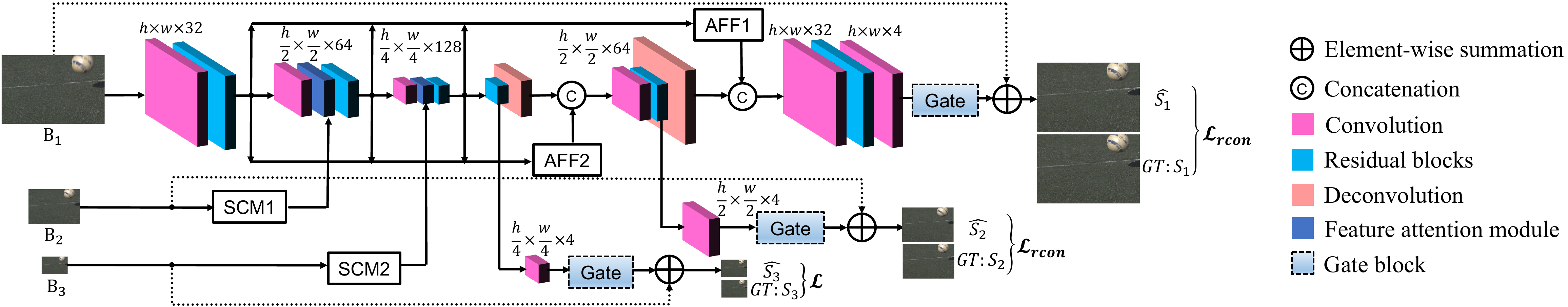}
    \caption{LBAG network: SCM and AFF denote shallow convolutional module asymmetric feature fusion module, respectively.}	
    \label{fig:network}
\end{figure*}

To bridge the gap between global and local deblurring, in this section, we introduce our local blur-aware gated deblurring model, LBAG, which detects blurred regions and restores locally blurred images simultaneously. Fig. \ref{fig:network} provides an overview of our proposed model. Deriving from MIMO-Unet \cite{cho2021rethinking}, we design a multi-scale Unet module. Different from global deblurring models, we specifically design a local blur detection method, gate blocks and blur-aware patch cropping strategy for local deblurring. 

\subsection{Ground-Truth Local Blur Mask Generation}
To supervise the training of blurred region detection, providing ground-truth local blur masks is necessary. We develop a Local Blur Foreground Mask Generator (LBFMG) to generate ground-truth local blur masks of training data, based on Gaussian Mixture-based background/foreground segmentation method \cite{zivkovic2004improved}, which outputs the foreground with the help of input backgrounds. In locally blurred images, all static regions are regarded as backgrounds. Capturing all the static regions separately in natural scenes is impracticable, because some moving objects are closely connected to the static items. To obtain the input background for a locally blurred image $\mathrm{B_T}$, we put all sharp and blurred images of the same scene except for $\mathrm{B_T}$ into LBFMG to update the background. Finally, $\mathrm{B_T}$ is put into LBFMG to generate its foreground mask as the ground-truth mask. The procedure of LBFMG is shown in Fig. \ref{fig:gate_module} and we put more details of it in our project's home page \ref{web}.

\subsection{LBAG: Local Blur-Aware Gated Network}
\begin{figure}
    \centering
    \includegraphics[width=0.48\textwidth]{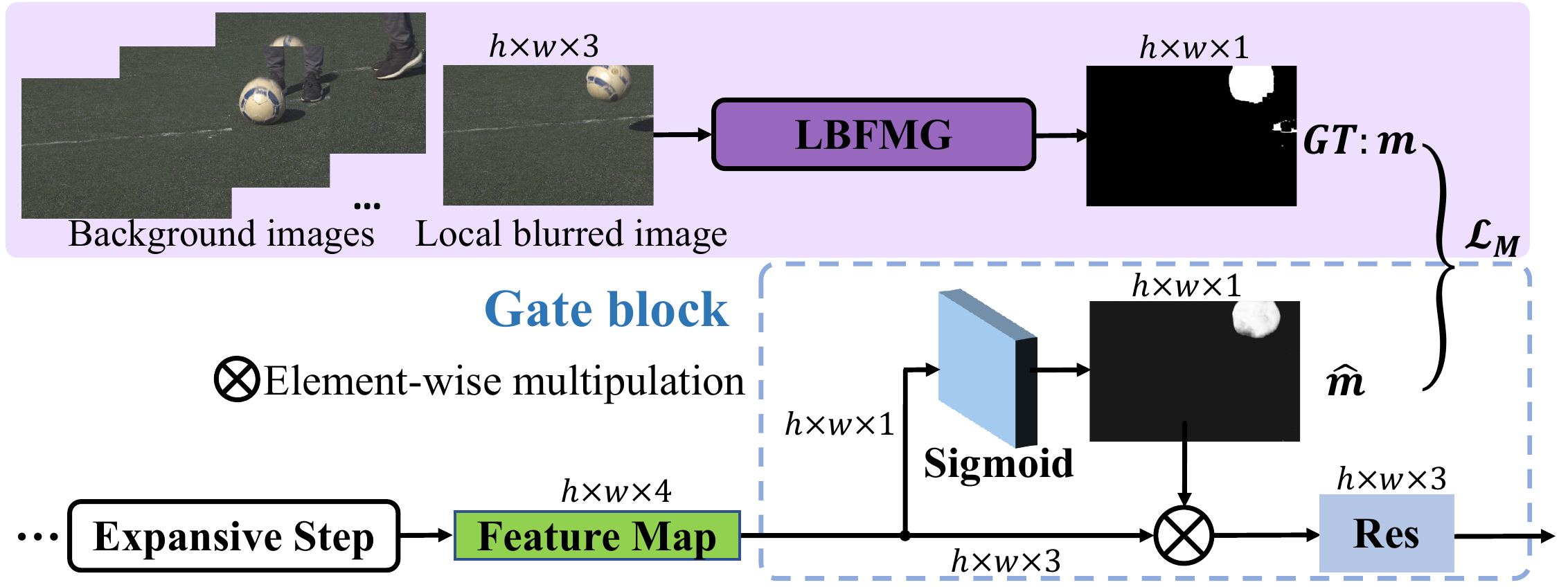}
    \caption{Ground-truth local blur mask generation (top) and the gate block (bottom).}	
    \label{fig:gate_module}
\end{figure}
We propose a local blur-aware gated network, LBAG, based on MIMO-Unet \cite{cho2021rethinking}. LBAG exploits the MIMO-Unet architecture as the backbone, and gate blocks at the end of the network, as shown in Fig. \ref{fig:network} and Fig. \ref{fig:gate_module}. The backbone consists of three contracting steps, three expansive steps, two shallow convolutional (SCM) modules and two asymmetric feature fusion (AFF) modules. We take a locally blurred image with a different scale as an input of every extracting step, as many CNN-based deblurring methods demonstrated that multi-scale images can better handle different levels of blur \cite{michaeli2014blind,liu2015blur,sun2015learning,nah2017deep,suin2020spatially}. To localize the local blurred regions, a gate block follows every expansive step.

As shown in Fig. \ref{fig:gate_module}, \textbf{the gate block} divides an input 4-channel feature map into a 3-channel latent and a 1-channel latent. The 1-channel latent passes through a sigmoid layer, forming a 1-channel pixel-level local blur mask prediction with pixel values ranging from 0 to 1, indicating how likely they are to be blurry (the higher the value, the more likely they are in locally blurred regions). For joint training, a mask-predicted loss is computed: $\mathcal{L}_M=MSE(m, \hat{m})$, where $m$ and $\hat{m}$ denote ground-truth local blur mask and predicted local blur mask. At the end of the gate structure, we multiply the 3-channel latent and the predicted local blur mask to compute the residual image of a locally blurred image and its corresponding sharp image. The gate block helps the network to localize the locally blurred regions, so that the global deblurring backbone could only modify pixels in the predicted locally blurred regions, without harming the static background of a local blurred image.

Finally, the multi-scaled residual images are added to their corresponding multi-scaled locally blurred images to form predicted sharp images.
A reconstruction loss $\mathcal{L}_{rcon}$ is calculated for supervised deblurring. 

$\mathcal{L}_{rcon}=\lambda_{2} \mathcal{L}_{ShM A E}+\lambda_{3} \mathcal{L}_{ShS S I M}+\lambda_{4} \mathcal{L}_{ShM S F R}$

$\mathcal{L}_{rcon}$ includes mean absolute error (MAE) loss $\mathcal{L}_{MAE}$, SSIM loss $\mathcal{L}_{SSIM}$ and multi-scale frequency reconstruction (MSFR) loss~\cite{cho2021rethinking} $\mathcal{L}_{MSFR}$. The total loss of LBAG is written as:
\begin{equation}\small
    \mathcal{L}=\lambda_{1} \mathcal{L}_{M}+\lambda_{2} \mathcal{L}_{M A E}+\lambda_{3} \mathcal{L}_{S S I M}+\lambda_{4} \mathcal{L}_{M S F R},
    \label{equ_loss}
\end{equation}
where $\lambda_1=0.01$, $\lambda_2=\lambda_3=1$, and $\lambda_4=0.1$. A shift-invariant operation is applied on the total loss for better image reconstruction, which is explained detailedly in our project's home page\ref{web}.

\subsection{Blur-Aware Patch Cropping Strategy}
In the local deblurring task, the local blurred regions only occupy a small percentage of the full image area (the average percentage is 11.75\% in ReLoBlur dataset). And thus a deep neural network can be easy to pay much more attention to the clear background than the blurred regions. To tackle this data imbalance problem, we develop a blur-aware patch cropping strategy (BAPC) in training. Specifically, we randomly crop 256$\times$256 patches from each training image with a 50\% chance; otherwise, we randomly select the patch center $ctr$ from the pixel in blurred regions marked by the ground-truth local blur mask, and then crop the patch centered with $ctr$. In this way, BAPC assures that LBAG can pay more attention to blurred regions and thus solves the image imbalance problem.

\section{Experiments and Analyses}

\subsection{Experiments}
\subsubsection{Experimental Settings}
LBAG is trained and evaluated on ReLoBlur dataset. We split the ReLoBlur dataset into 2010 pairs for training and 395 pairs for testing, without repeated scenes occurring in each split set.
Since the backbone of LBAG has the same objective as global deblurring networks, we can initialize the parameters of our backbone using the MIMO-UNet weights which are pre-trained on GoPRO dataset
for fast convergence. We denote the LBAG with pre-trained model initialization as LBAG+. We crop the images to 256$\times$256 patches as the training inputs using BAPC strategy. For data augmentation, each patch is horizontally or vertically flipped with a probability of 0.5. We use Adam \cite{kingma2014adam} as the optimizer, with a batchsize of 12 and an initial learning rate of 10$^{-4}$, which is halved every 100k steps. The training procedure takes approximately 70 hours (300k steps). For a fair comparison, we trained LBAG and the baseline deblurring methods for the same steps on 1 GeForce RTX 3090 with 24GB of memory. The model configuration of the baseline methods follows their origins.

In testing, we put locally blurred images with the full image size of 2152$\times$1436 into LBAG and baseline methods. We measure the full image restoration performances in terms of PSNR, SSIM. Because the ground-truth images are not strictly on the temporal centers of the moving tracks, we also calculate PSNR after temporally aligning the deblurred centers and the ground-truth centers of moving objects, which is denoted as PSNR$_a$, according to RealBlur \cite{rim2020real}. To evaluate local deblurring performances, we propose weighted PSNR and weighted SSIM:
\begin{equation}
\small
\begin{array}{lr}
    PSNR_{w}(\mathrm S,\hat{\mathrm S})=\frac{\sum_{x=1}^{N}\sum_{y=1}^{M}{\mathrm{PSNR}(\mathrm S(x,y), \hat{\mathrm S}(x,y))\times \mathrm{Msk(x,y)}}}{\sum_{x=1}^{N}\sum_{y=1}^{M}\mathrm{Msk(x,y)}}, \nonumber \\
    SSIM_{w}(\mathrm S,\hat{\mathrm S})=\frac{\sum_{x=1}^{N}\sum_{y=1}^{M}{\mathrm{SSIM}(\mathrm S(x,y), \hat{\mathrm S}(x,y))\times \mathrm{Msk(x,y)}}}{\sum_{x=1}^{N}\sum_{y=1}^{M}\mathrm{Msk(x,y)}}, \nonumber
\end{array}
\end{equation}
where $(x,y)$, $M$, $N$ and $Msk$ are pixel location, image width, image height, and the ground-truth local blur mask, respectively. $\mathrm S$ and $\hat{\mathrm S}$ denote sharp ground-truth and predicted sharp image, respectively.
\subsubsection{Local Deblurring Performance}
\label{Secion:deblurring_perf}
\begin{figure*}[t]
    \centering
    \includegraphics[width=\textwidth]{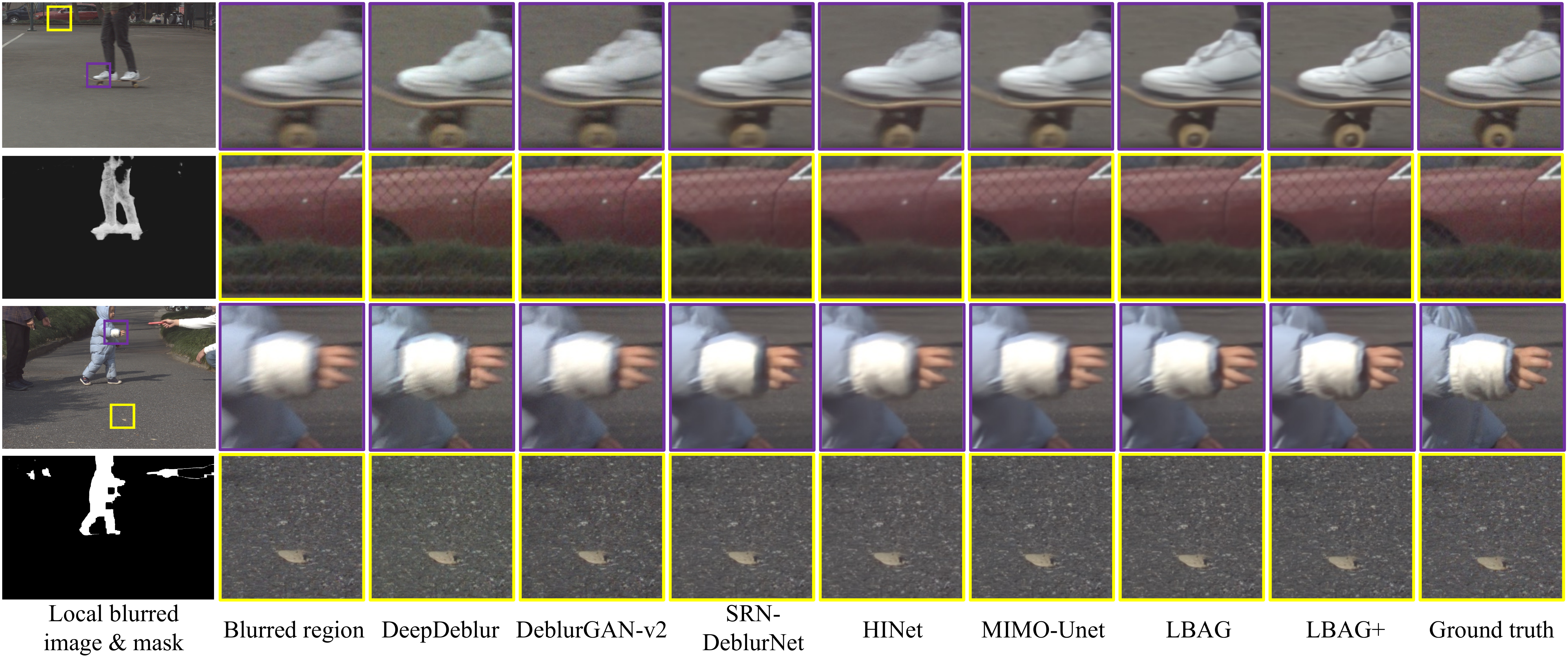}
    \caption{Visual comparison of different deblurring methods on ReLoBlur dataset: purple frames: locally blurred regions and deblurred regions; yellow frames: sharp regions and corresponding regions in deblurred images. More results are shown in our project's home page.}
    \label{fig:deblurred_results}
\end{figure*}

We compare the proposed method on ReLoBlur dataset with the following four deburring methods: DeepDeblur \cite{nah2017deep}, DeblurGAN-v2 \cite{Kupyn_2019_ICCV}, SRN-DeblurNet \cite{tao2018scale} and HINet \cite{chen2021hinet}. Fig. \ref{fig:deblurred_results} presents the visual comparison of deblurring results and local blur masks of a boy's hand and a man's shoe as instances. The results indicate that LBAG can well predict local blur regions and recover locally blurred images simultaneously. Compared with SOTA deblurring methods, LBAG removes the foreground motion blur without harming the textures, while SRN-DeblurNet deforms the foreground. For example, the hand and the shoe are distorted and lose consistency with the ground truths. Moreover, LBAG network generates sharper images with rich content, while DeepDeblur, DeblurGAN-v2, HINet and MIMO-Unet miss the detailed information. This demonstrates that global deblurring network is not suitable for local motion deblurring without focusing on the extreme blur changes of the foregrounds. Quantitative local deblurring results in Tab. \ref{table:resutls} show that LBAG exceeds other methods in terms of PSNR, SSIM, weighted PSNR and weighted SSIM with comparable model parameters. And with a pre-trained MIMO-Unet model, LBAG+ better reconstructs local motion-blurred images because it provides effective deblurring initialization. This proves that our methods are adept at removing local blur as well as maintaining the sharp background contents.
\setlength{\tabcolsep}{2pt}
\begin{table}[t]
\centering
\setlength{\tabcolsep}{0.5mm}
\begin{tabular}{c| c c c c c c}
\hline
Methods & $\uparrow$PSNR & $\uparrow$SSIM & $\uparrow$PSNR$_w$ & $\uparrow$SSIM$_w$ & $\uparrow$PSNR$_a$\\
\hline
DeepDeblur      & 33.05 & 0.8946  & 26.51 & 0.8152 & 33.70\\
DeblurGAN-v2    & 33.85 & 0.9027  & 27.37 & 0.8342 & 34.30\\
SRN-DeblurNet   & 34.30 & 0.9238  & 27.48 & 0.8570 & 34.88 \\
HINet           & 34.36 & 0.9151  & 27.64 & 0.8510 & 34.95\\
MIMO-Unet       & 34.52 & 0.9250  & 27.95 & 0.8650 & 35.42\\
\hline
LBAG  & 34.66 & 0.9249 & 28.25 & 0.8692 & 35.39\\
LBAG+ & \textbf{34.85} & \textbf{0.9257} & \textbf{28.32} & \textbf{0.8734} &\textbf{35.53}\\

\hline
\end{tabular}
\caption{Quantitative results of comparing local deblurring methods. ``PSNR$_w$'', ``SSIM$_w$'' and ``PSNR$_a$'' denote weighted PSNR, weighted SSIM and aligned PSNR, respectively. ``LBAG+'' denotes LBAG with pre-trained MIMO-Unet model.}
\label{table:resutls}
\end{table}

\subsection{Analyses}
\subsubsection{Dataset Analyses}
\begin{figure}[t]
     \centering   \includegraphics[width=0.48\textwidth]{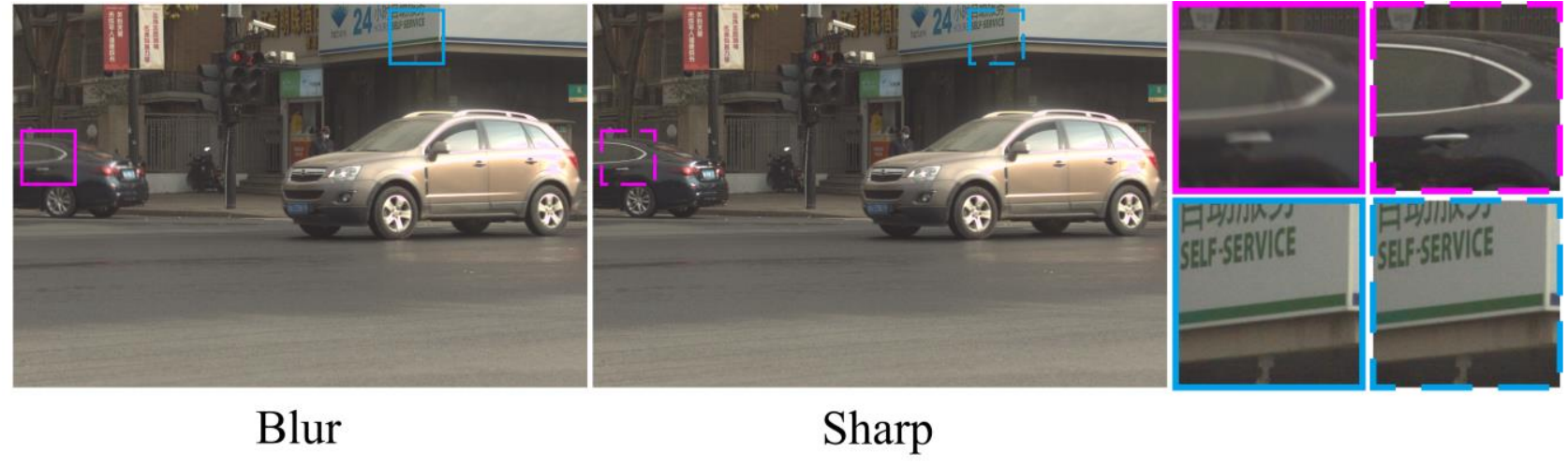}
     \caption{Examples of synthetic data: images in the 1$^{st}$ and 2$^{nd}$ columns are locally blurred images and their sharp ground-truths. The pink solid boxes and pink dotted boxes denote blurred regions from locally blurred images, and the corresponding sharp regions from sharp images. The blue solid boxes and the blue dotted boxes denote sharp regions from locally blurred images, and the corresponding sharp regions from sharp images.}
     \label{fig:synthetic_data}
\end{figure}
To evaluate the effectiveness and non-substitution of ReLoBlur dataset for local motion deblurring task, we train LBAG both on ReLoBlur dataset and synthetic local blurred data. Because there is no public local motion-blurred dataset, we construct synthetic data by adding local motion-blur to the sharp images of our ReLoBlur dataset. The construction of synthetic data is explained in our project's home page \ref{web}. We show an example of the synthetic data in Fig. \ref{fig:synthetic_data}.
\begin{figure}[h]
    \centering
    \includegraphics[width=0.48\textwidth]{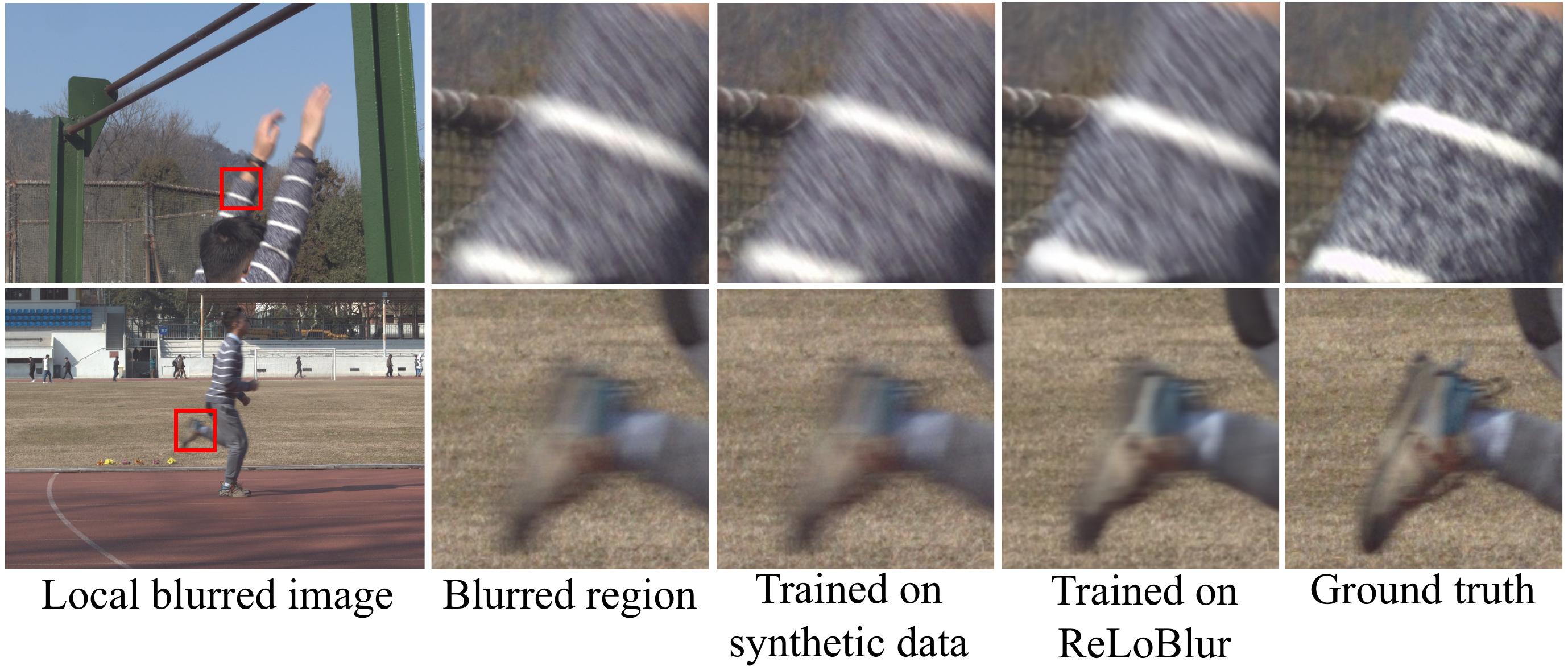}
    \caption{Visual deblurring results of synthetic-data-trained LBAG+ and ReLoBlur-trained LBAG+.}
    \label{fig:cmp_fake}
\end{figure}
\begin{table}[h]
\centering
\setlength{\tabcolsep}{1.8mm}
\begin{tabular}{c|cccc}
\hline
Training data & $\uparrow$PSNR & $\uparrow$SSIM & $\uparrow$PSNR$_w$ & $\uparrow$SSIM$_w$ \\
\hline
Synthetic data & 34.03 & 0.9015 & 27.42 & 0.8366 \\
ReLoBlur data & \textbf{34.85} & \textbf{0.9257} & \textbf{28.32} & \textbf{0.8734} \\
\hline
\end{tabular}
\caption{Quantitative results of synthetic-data-trained LBAG+ and ReLoBlur-trained LBAG+.}
\label{table:cmp_fake}
\end{table}

The deblurring results are shown in Fig. \ref{fig:cmp_fake} and Tab. \ref{table:cmp_fake}. We notice that the deblurring model trained on synthetic data could not remove the local blur and obtains lower quantitative scores on evaluation metrics. This demonstrates that synthetic blur remains a gap with real local blur, and the synthetic local blur data can not replace real local blurred  data for the training of local deblurring networks. Our ReLoBlur dataset enables efficient deep local deblurring training and vigorous evaluation compared with the synthetic data.

\subsubsection{Analyses of Local Deblurring Model}
\setlength{\tabcolsep}{0.1mm}
\begin{table}[h]
\centering
\resizebox{1.02\columnwidth}{!}{
\begin{tabular}{c|cccc|ccccc}
\hline
No. & Ga. & BAPC & $\mathcal{L}_{SSIM}$ & Pretr.&  $\uparrow$PSNR & $\uparrow$SSIM & $\uparrow$PSNR$_w$ & $\uparrow$SSIM$_w$ & $\uparrow$PSNR$_a$ \\
\hline
1(LBAG+) & $\checkmark$ & $\checkmark$ & $\checkmark$ & $\checkmark$ & \textbf{34.85} & \textbf{0.9257} & \textbf{28.32} & \textbf{0.8734} & \textbf{35.53} \\
2 &            & $\checkmark$ & $\checkmark$ & $\checkmark$ & 34.67 & 0.9256 & 27.88 & 0.8680 & 35.30\\
3 & $\checkmark$ &            & $\checkmark$ & $\checkmark$ & 34.75 & 0.9255 & 28.17 & 0.8695 & 35.48 \\
4 & $\checkmark$ & $\checkmark$ &            & $\checkmark$ & 34.62 & 0.9254 & 27.85 & 0.8639 & 35.28\\
5(LBAG) & $\checkmark$ & $\checkmark$ & $\checkmark$ &      & 34.68 & 0.9256 & 28.10 & 0.8677 & 35.43\\
\hline
\end{tabular}
}
\caption{Ablations of LBAG on ReLoBlur. \textit{Ga.} and \textit{Pretr.} denote abbreviations of gate block and pre-training strategy.}
\label{table:ablation_study}
\end{table}

To verify the effects of our proposed local deblurring techniques, we compare LBAG with or without gate blocks, BAPC strategy, SSIM loss term and pre-trained model parameter initialization, as shown in Tab. \ref{table:ablation_study}. We have the following observations:
\begin{itemize}
\item Comparing line 1 and line 2, we can see that both global metrics (weighted PSNR and weighted SSIM) and local metrics (PSNR and SSIM) drop when removing the gate blocks, indicating that gate blocks can effectively neglect background objects and focus on blurry objects.
\item Comparing line 1 and line 3, we find that both global and local metrics drop without BAPC strategy, which proves BAPC's importance. Besides, the local metrics drop greater than global metrics, which further demonstrates that BAPC strategy encourages our model to pay more attention to blurred regions and facilitates training.
\item Comparing line 1 and line 4, we see that SSIM loss term can significantly improve the SSIM-related scores.
\item Comparing line 1 and line 5, we notice that the absence of loading pre-trained model limits network expressions, because the pre-trained MIMO-Unet model produces effective deblurring initialization for our gated network.
\end{itemize}

\section{Conclusion}
This paper dealt with local blur problem and bridged the gap between global and local deblurring tasks. We constructed the first real local motion blur dataset, ReLoBlur, containing sharp and locally blurred images in real scenes. We addressed color cast and misalignment problems occurred in original-shot images through our novel post-processing pipeline. To conquer the challenges in local deblurring, we proposed a local blur-aware gated method, LBAG, with local deblurring techniques including a local blur region detection method, gate blocks and a blur-aware patch cropping strategy. Extensive experiments show that ReLoBlur enables efficient deep local deblurring and vigorous evaluation, and LBAG outperforms the SOTA local deblurring methods qualitatively and quantitatively. In the future, we will fasten LBAG inference speed and improve the deblurred image quality by introducing generative models.

\section*{Acknowledgments}
This work is supported by Jiangsu Science and technology development Foundation.

\section{Appendix}
\subsection{A. Color Cast Problems in RealBlur Dataset}
\label{appendix:previous_dataset}
\begin{figure}[h!]
    \centering
    \includegraphics[width = 0.48\textwidth]{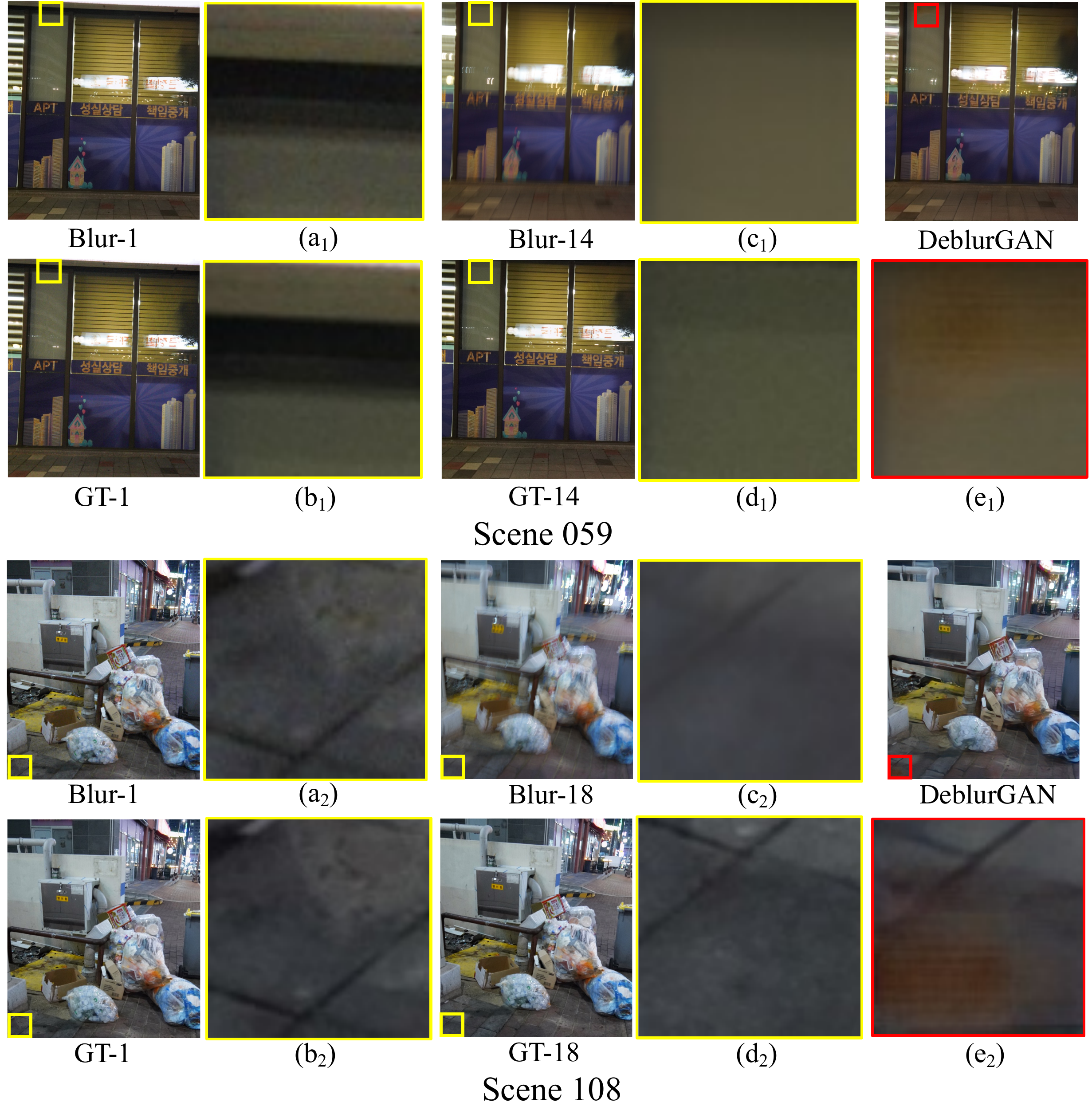}
    \caption{Color cast problems occurred in RealBlur dataset and how they influence deblurring neural network's training results: (a$_1$), (a$_2$), (b$_1$) and (b$_2$) are patches of reference image pairs. (c$_1$), (c$_2$), (d$_1$) and (d$_2$) are patches of blur-sharp image pairs. (e$_1$) and (e$_2$) are deblurring results of DeblurGAN on RealBlur-J. The two examples are from Scene 059 and Scene 108 in RealBlur-J.}
    \label{fig:RealBlur}
\end{figure}

 RealBlur dataset \cite{rim2020real} contains local degradations of color casts in paired images. The colors of the same area in a pair of sharp-blur images are different. This should be avoided because color casts may reduce the training performance in local deblurring tasks. Fig. \ref{fig:RealBlur} provides examples of the color cast problem in RealBlur-J. In Scene 059, both the reference image pair (Blur-1 and GT-1) and the blur-sharp pair (Blur-14 and GT-14) contain color cast locally. Blur-1's window frame area contains more colorful stripes than GT-1's window frame area does. The Blur-14's window area contains more orange elements than the GT-14's window area does. The same problems occur in Scene 108. After deblurring by DeblurGAN on RealBlur dataset, color casts are more pronounced in the deblurring results. Different from RealBlur dataset, we apply color correction and photometrical alignment in post-precessing in Section 3 (in the main body of our paper). Fig. \ref{fig:whiteboard} and Fig. \ref{fig:colorboard} in Appendix D \ref{section:analyse_postp} prove that our effort is well-assured.

\subsection{B. The Authenticity of ReLoBlur Dataset}\label{section:analyse_postp}
We prove the authenticity of ReLoBlur dataset by quantitatively and perceptually evaluating the results of our paired image post-processing pipeline, including color correction, photometrical alignments and geometrical alignments.

\begin{figure}[!h]
	\centering
	\begin{subfigure}[h]{0.48\textwidth}
		\centering
		\includegraphics[width=\textwidth]{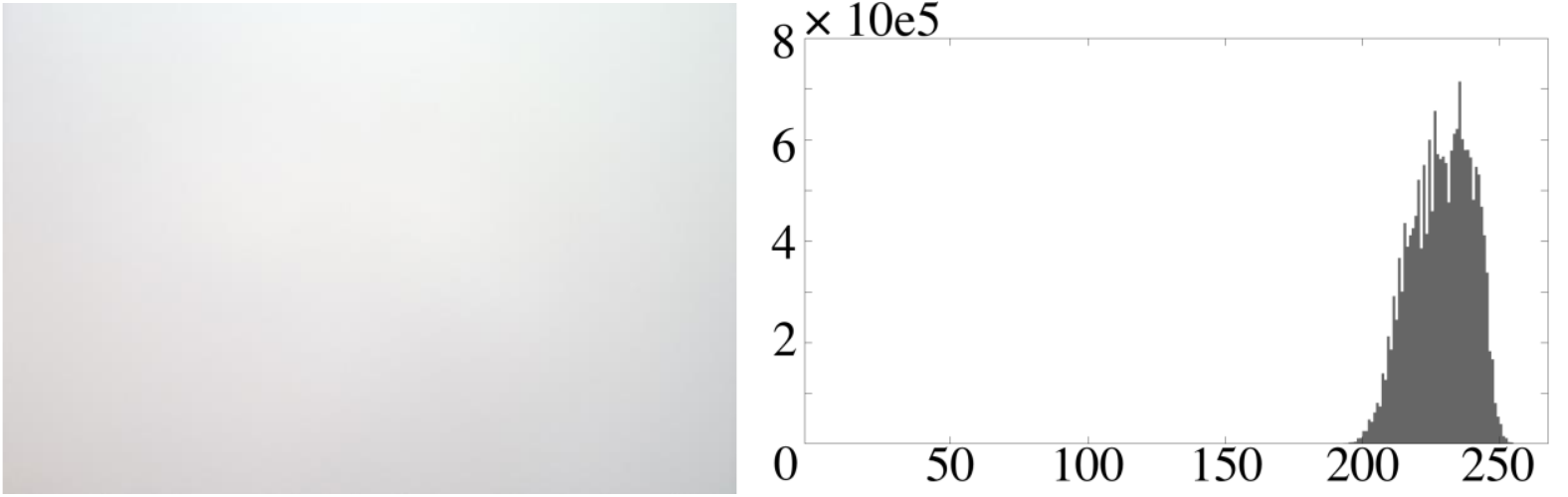}
    	\caption{before color correction.}
	\end{subfigure}
	\begin{subfigure}[h]{0.48\textwidth}
		\centering
		\includegraphics[width=\textwidth]{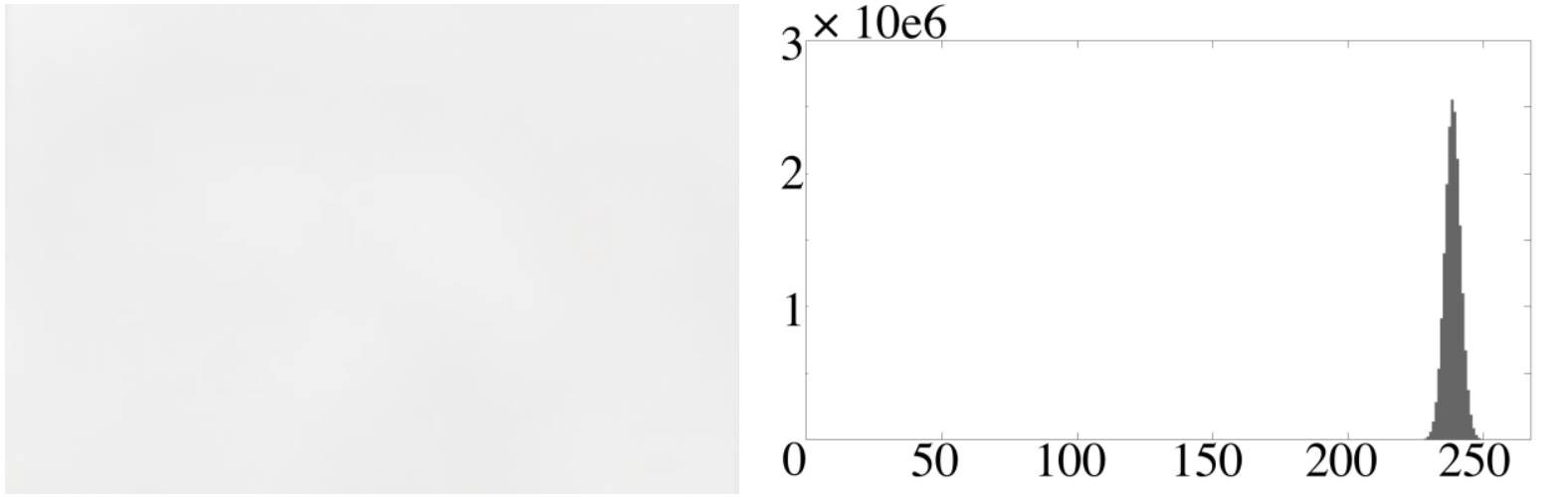}
    	\caption{after color correction.}
	\end{subfigure}
    \caption{Color correction results. The left and right images in each subfigure are whiteboard photos and histograms.}	
    \label{fig:whiteboard}
\end{figure}

To evaluate the color correction performances, we capture a standard whiteboard in a standard 6500K reflective lightbox. As shown in Fig. \ref{fig:whiteboard}, before color correction, there exists non-uniform color cast in each camera and the grayscale values disperse from 200 to 255. After color correction, the photo of the whiteboard presents uniform white color and the grayscale values are more concentrated (230$\sim$250). 

\begin{figure}[!h]
    \centering
    \includegraphics[width=0.47\textwidth]{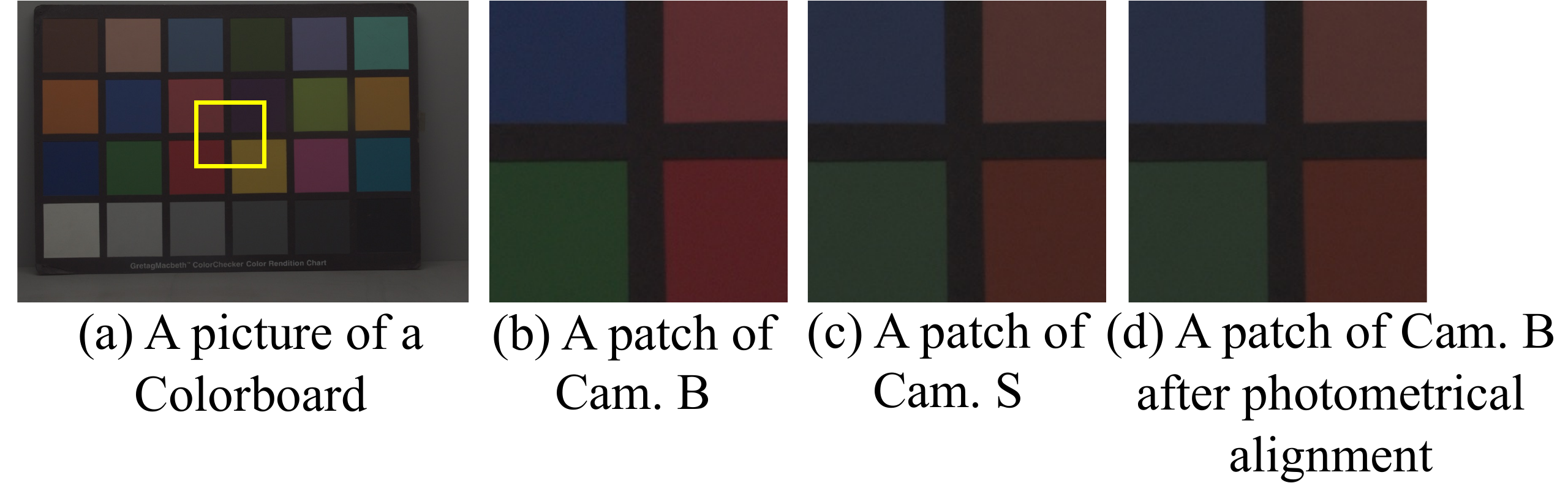}
    \caption{Phometrical alignment results.}
    \label{fig:colorboard}
\end{figure}

Results of photometrical alignment are depicted in Fig. \ref{fig:colorboard}, which shows that there exists an apparent brightness difference between the original-shot photo captured by Camera B and Camera S. After photometrical alignment, the two photos share almost the same brightness and color.

We quantitatively evaluate the accuracy of color correction and photometric alignment by computing a $\Delta \mathrm{L}$ function by pixels between the static sharp background ($l_1$) and blur background ($l_2$), which is defined as:

\begin{equation}\small
    \small
    \Delta \mathrm{L}= \frac{\sum_{x=1}^{M} \sum_{y=1}^{N}\left| l_{1}(x,y)-l_{2}(x,y)\right|}{\sum_{x=1}^{M} \sum_{y=1}^{N}l_{2}(x,y)},
\end{equation}
where $M=2152$ and $N=1436$ are the image width and height, respectively. $(x,y)$ denotes pixel location. The average $\Delta \mathrm{L}$ of the static image pairs of all scenes is 2.870\%, which implies that there is no obvious color cast or brightness difference perceptually. 

\begin{figure}[!h]
    \centering
    \includegraphics[width=0.48\textwidth]{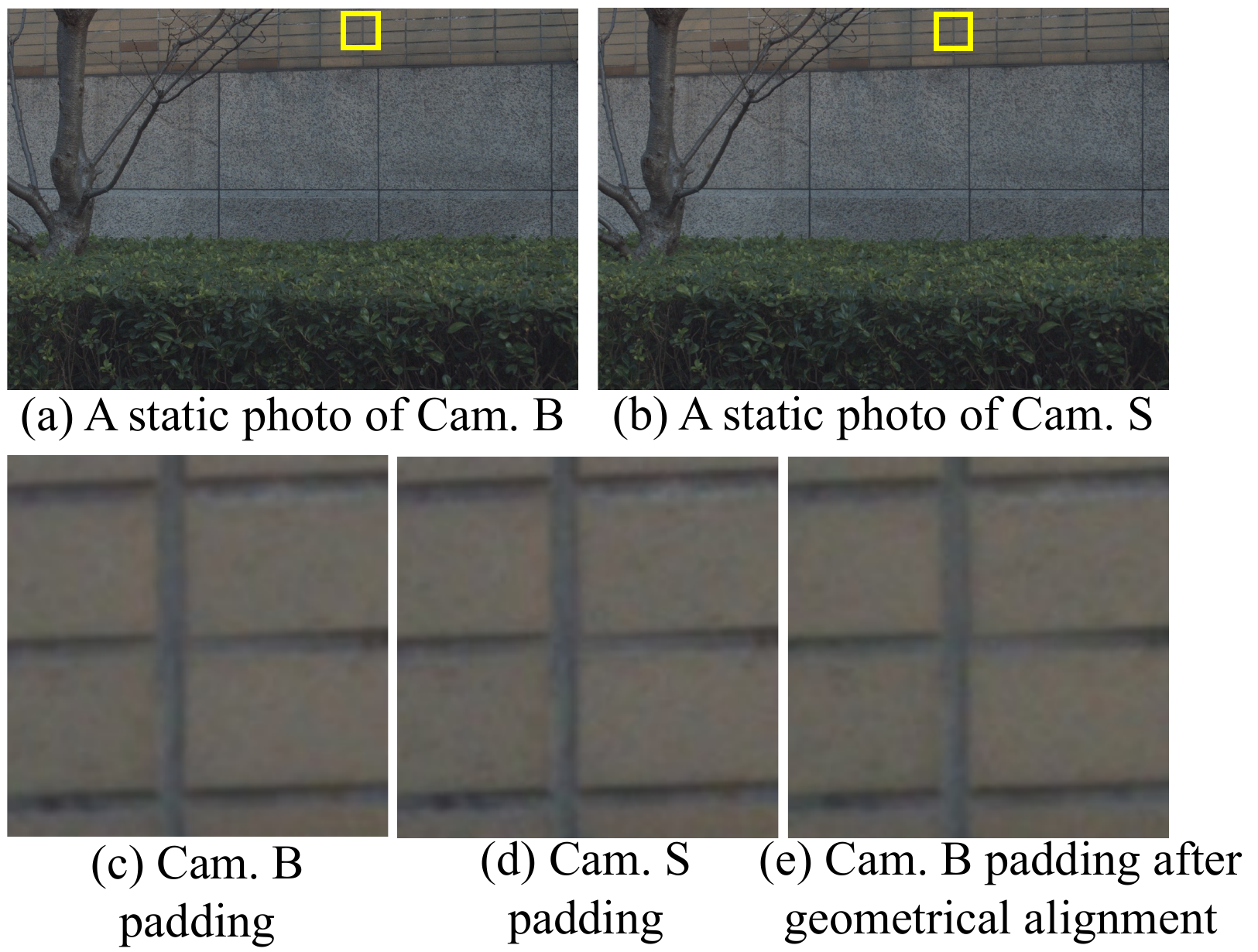}
    \caption{Geometrical alignment results.}
    \label{fig:static}
\end{figure}

For geometrical alignment evaluation, we compare the PSNR of the static background image pairs before and after geometrical alignment. Fig. \ref{fig:static} shows that there is no obvious misalignment between the sharp and locally blurred image perceptually. As illustrated in Tab. \ref{table:post_p}, the average PSNR of the static aligned image pairs is 36.79dB, which is 10dB over that of the original-shot image pairs. Both the visual results and quantitative results above prove that our paired image post-processing pipeline is well-assured.
\begin{table}[!h]
\small
\setlength{\tabcolsep}{1mm}
\centering

\caption{Quantitive results of image post-processing}
\label{table:post_p}

\begin{tabular}{c|cccc}
\hline
Post-processing & $\Delta{L}$ & $\uparrow$PSNR & $\uparrow$SSIM & \makecell[c]{Geometrical\\Error (pixels)} \\
\hline
Before & 4.550\% & 26.14 & 0.5859 & 3 \\
After  & 2.870\% & 36.79 & 0.9091 & 1 \\
\hline
\end{tabular}
\end{table}
\setlength{\tabcolsep}{1.4pt}

\subsection{C. LBFMG Algorithm}
We propose a local blur foreground mask generator, LBFMG, to generate the ground truths for the training of local blur mask prediction. LBFMG applies a background substractor, \textit{BackgroundSubtractorMOG2}\footnote{A function based on Gaussian Mixture-based Background/Foreground Segmentation Algorithm, which can be viewed on https://docs.opencv.org/3.4/d7/d7b/classcv\_1\_1BackgroundSubtractorMOG2.html} to extract the foreground of an image. For the initialization of the background, we utilize static images, which only contain the sharp background without moving objects in the scene. For each local blur mask $Msk$, the generation procedure is shown in Fig. 5 in the main body of our paper. And more generating details are illustrated in Algorithm \ref{alg: LBFMG}.
\begin{algorithm}
    \centering 
    \caption{LBFMG}\label{alg: LBFMG} 
    \begin{algorithmic}[1] 
    \STATE\textbf{Load}: \textit{BackgroundSubtractorMOG2} function \textit{G}, a zero matrix M with the size of 2152$\times$1436.
    \STATE\textbf{Input}: local blurred image $\mathrm{B}_1$ and its corresponding sharp image $\mathrm{S}_1$, static sharp image $\mathrm{S}_0$ and its corresponding static image $\mathrm{B}_0$, other image pairs $P=\{(\mathrm{S}_k,~\mathrm{B}_k)\}_{k=2}^{K-1}~(K \mathrm{: total~number~of~image~pairs})$ of the scene.
    \STATE\textbf{Initialize}: put the static image $\mathrm{S}_0$ and $\mathrm{B}_0$ into \textit{BackgroundSubtractorMOG2} to initialize the sharp foreground, $fgmaskS$, and the blur foreground, $fgmaskB$, respectively.
    \REPEAT 
        \STATE Sample $k \sim \{2,3,...,K\}$ in order
        \STATE Put a sharp image $\mathrm{S}_k \sim P$ into $\textit G_1$ to update $fgmaskS$
        \STATE Put a local blurred image $\mathrm{B}_k \sim P$ into $\textit G_2 $ to update $fgmaskB$
    \UNTIL{$k=K$} 
    \STATE Put the sharp image $\mathrm{S}_1$ into $\textit G_1$ to generate $fgmaskS$.
    \STATE Put the local blurred image $\mathrm{B}_1$ into $\textit G_2$ to generate $fgmaskB$.
    \STATE Calculate M by Equation: M = $(fgmaskS\textgreater1)$ OR $(fgmaskB\textgreater1)$.
    \STATE Generate the ground-truth mask of local blurred image $\mathrm{B}_1$ with image erosion and dilation.
    \STATE\textbf{Output}: Ground-truth mask $Msk_1$.
    \end{algorithmic} 
\end{algorithm} 

\begin{figure*}[t]%
    \centering
    \includegraphics[width=\textwidth]{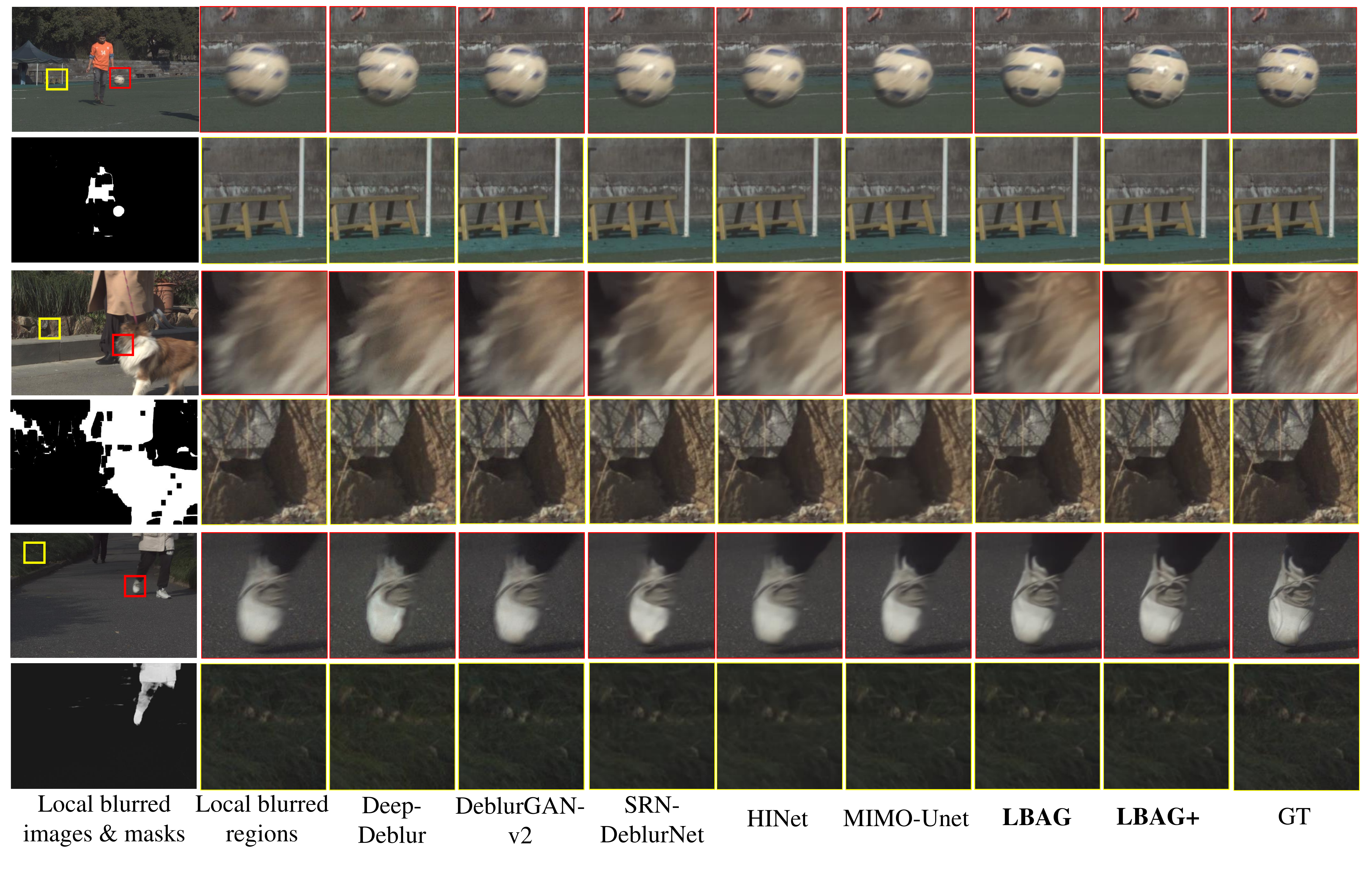}
    \caption{More visual comparisons of different deblurring methods on ReLoBlur dataset: red frames denote locally blurred regions and deblurred regions; yellow frames denote sharp regions and the corresponding regions in deblurred images. The $1^{st}$ pictures in the $2^{nd}$, $4^{th}$ and 6$^{th}$ row are predicted blur masks by our proposed LBAG.}
    \label{fig:more_deblurring_results}
\end{figure*}

\subsection{D. Losses and Shift-Invariant Operation}
\begin{figure}[h!]
     \centering
     \includegraphics[width=0.4\textwidth]{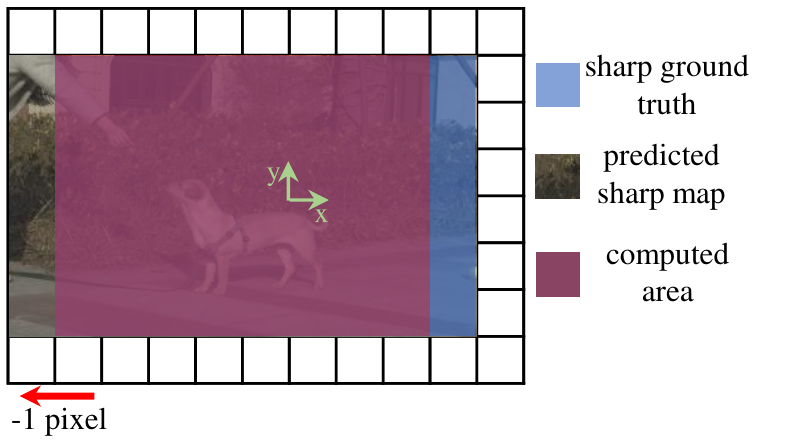}
     \caption{Shift-invariant operation.}
     \label{fig:shift}
\end{figure}

To compensate for the geometrical alignment error (about 1 pixel), we apply a shift-invariant operation when computing the total loss function (Equ. 7 in the main body of our paper), including a mask predicted loss $\mathcal{L}_{M}$, and a reconstruction loss $\mathcal{L}_{rcon}$. The shift-invariant operation moves a predicted sharp map $\hat{\mathrm S}=\mathcal{M}(\mathrm B)$ for (k,l) pixel, $k,l\in\{-1,0,1\}$, where $k$ and $l$ denote the moving pixel towards $x$ and $y$ direction, respectively. ``-'' means the opposite axis. $\mathcal{M}$ means the proposed local motion deblurring model. $\mathrm{B}$ means a locally blurred image. As shown in Fig. \ref{fig:shift}, we compute the total loss functions on the overlapping area of the ground-truth sharp image $\mathrm S$ (or the ground-truth mask $\mathrm m$) and the shift-invariant predicted sharp map $\hat{\mathrm S}$ (or the predicted local blur mask $\hat{\mathrm m}$). With the shift-invariant operation, the total loss can be also written as:
\newcommand{\ShiftOp}{\mathrm{Shift}}
\newcommand{\Msk}{\mathrm{Msk}}
\newcommand{\MAE}{\mathrm{MAE}}
\newcommand{\SSIM}{\mathrm{SSIM}}
\newcommand{\MSE}{\mathrm{MSE}}
\newcommand{\LShMAE}{\mathcal{L_{\mathrm{ShMAE}}}}
\newcommand{\LSSIM}{\mathcal{L}_{\mathrm{SSIM}}}
\newcommand{\LShSSIM}{\mathcal{L_{\mathrm{ShSSIM}}}}
\newcommand{\LShMSFR}{\mathcal{L_{\mathrm{ShMSFR}}}}
\newcommand{\LShM}{\mathcal{L}_{\mathrm{ShM}}}
\newcommand{\LMSFR}{\mathcal{L}_{\mathrm{MSFR}}}
\begin{equation}\small
    \begin{aligned}
        \mathcal L_{Sh}(\hat{\mathrm S},\mathrm S) =
        &\lambda_1\LShM(\hat{\mathrm m},\mathrm m) + \lambda_2\LShMAE(\hat{\mathrm S},\mathrm S)\\
        &+\lambda_3\LShSSIM(\hat{\mathrm S}, \mathrm S)
        +\lambda_4\LShMSFR(\hat{\mathrm S},\mathrm S),
    \end{aligned}
 	\label{equ_total_loss}
 \end{equation}
 where $\lambda_1=0.01, \lambda_2=\lambda_3=1, \lambda_4=0.1$ are constant loss weights. The shift-invariant MSE loss, MAE loss, SSIM loss and MSFR loss follow the derivation below:
 \begin{equation}\small
     \LShM(m,\hat{\mathrm m}) = \min_{\forall k \in \{-1,0,1\}, \atop l \in \{-1,0,1\}}(\MSE(\mathrm m, \ShiftOp_{k,l}(\hat{\mathrm m}))),
 	\label{equ_LShM}
 \end{equation}
 \begin{equation}\small
     \LShMAE(S,\hat{\mathrm S}) = \min_{\forall k \in \{-1,0,1\}, \atop l \in \{-1,0,1\}}(\MAE(\mathrm S, \ShiftOp_{k,l}(\hat{\mathrm S}))),
 	\label{equ_LShMAE}
 \end{equation}
\begin{equation}\small
    \left\{
        \begin{aligned}
            &\LSSIM(\mathrm S,\hat{\mathrm S}) =-\frac{1}{N*M}\sum_{x=1}^{N}\sum_{y=1}^{M}{\SSIM(\mathrm S(x,y), \hat{\mathrm S}(x,y))},\\
            &\LShSSIM(\mathrm S,\hat{\mathrm S})= \min_{\forall k \in \{-1,0,1\}, \atop l \in \{-1,0,1\}}(\LSSIM(\mathrm S, \ShiftOp_{k,l}(\hat{\mathrm S}))), 
         \end{aligned}
    \right.
\label{equ_SSIM}
\end{equation}
\begin{equation}\small
	    \left\{
 	        \begin{aligned}
                 &\LMSFR(\mathrm S,\hat{\mathrm S})=\sum_{k=1}^{K} \frac{1}{t_{k}}\left\|\mathcal{F}\left(S_{k}\right)-\mathcal{F}\left(\hat{S}_{k}\right)\right\|_{1},\\
                 &\LShMSFR(\mathrm S,\hat{\mathrm S})= \min_{\forall k \in \{-1,0,1\},\atop l \in \{-1,0,1\}}(\LMSFR(\mathrm S, \ShiftOp_{k,l}(\hat{\mathrm S}))).  
	        \end{aligned}
         \right.
 	\label{equ_LSSIM}
\end{equation}

\subsection{E. More Local Deblurring Results}
We provide more local deblurring results of comparing the methods on ReLoBlur datasets, as shown in Fig. \ref{fig:more_deblurring_results}.

\subsection{F. The Construction of Synthetic Data}
Firstly, we select the foreground regions using COCO-Instance Segmentation model\footnote{https://github.com/facebookresearch/detectron2} \cite{he2017mask}. Then, we use \textit{Kornia} filter \cite{zhang2019making} to blur the selected foreground regions with a $5\times5$ blur kernel and different moving modes, to simulate local motion blur. The moving modes include translation and rotation. 

\bibliography{aaai23}

\end{document}